\def\year{2019}\relax
\documentclass[letterpaper]{article} %DO NOT CHANGE THIS
\usepackage{aaai19}  %Required
\usepackage{times}  %Required
\usepackage{helvet}  %Required
\usepackage{courier}  %Required
\usepackage{url}  %Required
\usepackage{subfigure}
\usepackage{multirow}
\usepackage{booktabs}
\usepackage{caption}
\usepackage{graphicx}  %Required
\begin{document}
\title{Distant Supervision for Relation Extraction with Linear Attenuation\\Simulation and Non-IID Relevance Embedding}

\author{Changsen Yuan\textsuperscript{1}, Heyan Huang\textsuperscript{1,2\thanks{Corresponding author}}, Chong Feng\textsuperscript{1}, Xiao Liu\textsuperscript{1}, Xiaochi Wei\textsuperscript{3}\\
1.Department of Computer Science, Beijing Institute of Technology, China\\
2.Beijing Engineering Research Center of High Volume Language Information Processing \\and Cloud Computing Applications, China\\
3.Baidu Inc. , China\\
\{yuanchangsen, hhy63, fengchong, xiaoliu\}@bit.edu.cn weixiaochi@baidu.com\\
}

\maketitle
\begin{abstract}

%修改However, there are some disadvantages on selecting important words in the sentence and capturing relevance of sentences in the bag when using the existing methods to extract relations. In this paper, we propose a novel model to respectively address these problems.

Distant supervision for relation extraction is an efficient method to reduce labor costs and has been widely used to seek novel relational facts in large corpora, which can be identified as a multi-instance multi-label problem.
However, existing distant supervision methods suffer from selecting important words in the sentence and extracting valid sentences in the bag.
%However, there are some imperfections on selecting important words in the sentence and capturing the relevance of sentences in the bag when applying the existing methods to extract relations.
Towards this end, we propose a novel approach to address these problems in this paper.
Firstly, we propose a linear attenuation simulation to reflect the importance of words in the sentence with respect to the distances between entities and words.
Secondly, we propose a non-independent and identically distributed (non-IID) relevance embedding to capture the relevance of sentences in the bag.
Our method can not only capture complex information of words about hidden relations, but also express the mutual information of instances in the bag.
%We conduct extensive experiments on a widely used dataset and the experimental results show that our method significantly outperforms all baseline systems.
Extensive experiments on a benchmark dataset have well-validated the effectiveness of the proposed method.

%To solve the decay word, we assume that when the distance increases between entities and words in the sentence, the connection of entities and words perform linearity decrease. To address the relevance of sentences in the bag, we use the concept of non-independent and identically distributed (non-IID) to express the spatially-connected or temporally-connected with the adjacent sentences. Our method not only capture more information of words for predicting relations, but also express mutual information of each instances in the bag. We conduct extensive experiments on a widely used dataset and the experimental results show that our method outperforms all baseline systems significantly.

\end{abstract}

\section{Introduction}
Relation extraction, aiming to categorize semantic relations between entity pairs in plain texts, has been widely adopted in many natural language processing (NLP) tasks, such as question answering \cite{DBLP:conf/cvpr/SadeghiDF15}, text categorization \cite{DBLP:conf/acsc/HuynhTMS11} and web search \cite{DBLP:conf/acl/YanOMYI09}. Traditional supervised methods for relation extraction require a large amount of high-quality corpus for model training, which is extremely expensive and time-consuming. Additionally, these datasets are often restricted to certain domains.
%In recent years, distant supervision for relation extraction which can automatically find abundant relational facts has been proposed to work out this problem.
In recent years, distant supervision for relation extraction has been proposed to find abundant relational facts with large amount auto-generated labels.
However, it has two major flaws in existing distant supervision methods.

Firstly, the existing approaches acquiescently assume that each word in the sentence has the same weight in relation extraction. This hypothesis is too strong and usually leads to wrong labels.
%With the increase of distance between entities and words, words can not keep the same weight in distant supervision.
%As the distances between entities and words increase, their relationships are gradually weak. Therefore, words can not maintain the same weight in distant supervision.
The relationship between entities and words gradually decreases with the creasing of the distant between them. Therefore, words can not maintain the same weight in distant supervision.
%For example, there are approximately half of sentences longer than 40 words in the New York Times corpus (NYT)\footnote{http://iesl.cs.umass.edu/riedel/ecml.}. \cite{Mcdonald2007Characterizing} showed that the accuracy of syntactic parsing decreases significantly with increasing sentence length. And we also can clearly observe that in Figure 1, \emph{businessman} and \emph{politician} have different weights about \emph{Trump} in the S2, which are 0.85 and 0.77 respectively.
For example, (\emph{South Korea}, \emph{Seoul}, \emph{Country}) is a relational fact in KB. Each word in the sentence ``\emph{many foreign investors say the investigation is emblematic of the political uncertainty they face in investing in South Korea, a concern that looms large as Washington and Seoul are negotiating a free trade agreement.}" is not always useful for ``\emph{Country}". Some invalid words exist in the long sentence. Moreover, \cite{Mcdonald2007Characterizing} showed that the accuracy of syntactic parsing decreases significantly with increasing sentence length.
In the bag, we find that some instances are too long and contain some invalid words about target relation. And these invalid words are usually far away from entities.
Long distance between entity and word indicates a weak correlation between them. Conversely, short distance between entity and word possesses strong correlation. These phenomena sometimes lead to wrong labels in distant supervision. Therefore, if we use the same weights about words in relation extraction, weights of words will not only affect the expression of sentences, but also have an important impact on the judgement of labels.

\begin{figure*}
  \centering
  \includegraphics[scale=0.3]{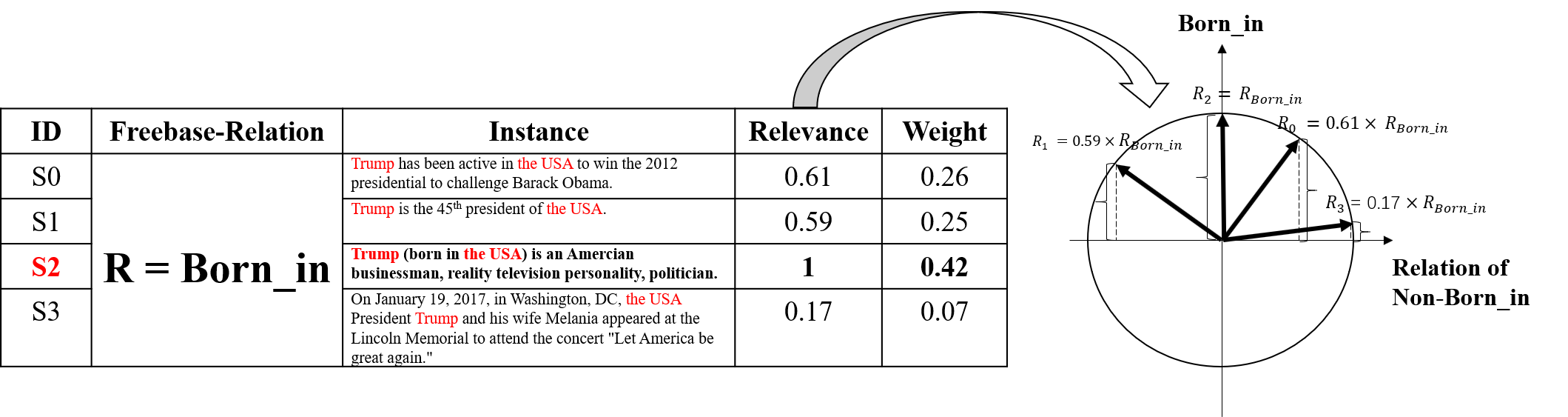}
  \caption{An example of non-IID relevance embedding in a bag. There are 4 sentences and 3-rd sentence is the best sentence to express relation of ``\emph{Born\_in}". The right of Figure 1 represent the relevance of sentences. The vertical axis represents relation of ``\emph{Born\_in}" with coordinating value from -1 to 1, while other directions represent other relations.}
\end{figure*}

Secondly, distant supervision for relation extraction possesses an ideal hypothesis that all instances containing the same entity pairs express the same relation. However, this is far from reality, because there may exist multi relations between a specific entity pairs. For example, both the relation ``\emph{Born\_in}" and ``\emph{Employ\_by}" are valid between the entity pair ``\emph{Trump}" and ``\emph{the USA}".
%this hypothesis sometimes produces some invalid sentences. Because it is possible that these sentences may not belong to the same relation and they simply share the same topic.
To solve this problem, the multi-instance learning \cite{Hoffmann2011Knowledge,DBLP:conf/emnlp/SurdeanuTNM12} and sentence-level attention \cite{Lin2016Neural,Ji2017Distant} have been proposed, but the above approaches also have flaws.
In relation extraction, the multi-instance learning only selects the instance with the highest probability to be a valid candidate, so that a large amount of rich information is lost. And the sentence-level attention considers instances in the bag as independent and identically distributed (IID), therefore, the relevance of instances is ignored consequently.
In contrast, these instances with the same entity pairs in the bag have more or less connections, which are important information of sentences. Toward this end, we assume that the relevance of sentences is able to selectively assign higher weights for valid sentences and lower weights for invalid sentences.
For example, in Figure 1, sentence S1 expresses the relation ``\emph{Employ\_by}" and sentence S2 expresses the relation ``\emph{Born\_in}". But we can implicitly obtain the relation of ``\emph{Born\_in}" between ``\emph{Trumph}" and ``\emph{the USA}" from the S1. This phenomenon illustrates that there is the connection between two sentences.
Therefore, non-independent and identically distributed (non-IID) are proposed to solve the relevance of instances and enhance valid sentences.

In this paper, we propose linear attenuation simulation and non-IID relevance embedding to increase valid instances and enhance the results of relation extraction. To address the first problem, we assume that the connection of entity and word changes with the distance between entity and word. This variation is linear attenuation.
Linear attenuation simulation can reduce the weight of word with the increase of distance between entity and word. Thus, we use linear attenuation simulation to work out this problem.

To solve the next problem, we adopt non-IID relevance embedding to learn the relevance of instances. Non-IID relevance embedding builds non-IID representations via modeling each bag along with its corresponding neighbors. Concretely, we use the cosine similarity between two sentences ($S_1$, $S_2$) to represent the relevance of $S_2$ about $S_1$, where $S_1$ is the best sentence to express the relation. If the sentence has lower similarity with the best sentence which can perfectly perform relation in the bag, this sentence will be assigned to a low weight.
%Our method not only can capture the relevance of instances, but also can alleviate the wrong labels for relation extraction.
%Therefore, the weights of valid sentences are improved via non-IID relevance embedding, and then enhance the correct labels for relation extraction.
Therefore, the non-IID relevance embedding can improve the weights of valid sentences and enhance the correct labels for relation extraction.
The experimental results show that our method achieves significant and consistent improvements in relation extraction as compared with the state-of-the art methods.

The main contributions of this paper are summarized as follows:
\begin{itemize}

%\item We explore connections between each word and entities and propose word attention to select useful words in relation extraction. And word attention is proposed to solve this problem.
\item We propose a linear attenuation simulation to select useful words and alleviate the wrong labels which are caused by long distance between entity and word.

\item To address the relevance of sentences, we develop innovative solutions that introduce non-IID relevance embedding to distant supervised relation extraction.

\item In the experiments, results show that our model achieves better performance in distant supervised relation extraction.

\end{itemize}

\section{Methodology}
We propose a new model for relation extraction containing linear attenuation simulation and non-IID relevance embedding. Linear attenuation simulation not only can provide and remain important words, but also improve the representation of sentences in our model. Non-IID relevance embedding provides more information between each sentence in the bag, which is able to select valid instances and bring more relevant information.
The overall structure of our proposed model is illustrated in Figure 2, our model consists of two main components: PCNNs Module and Attention Module.
The PCNNs Module is used to extract features and compute the weights of words from a sentence in a bag. And the PCNNs Module is further comprised of \emph{Vector Representation}, \emph{Linear Attenuation Simulation}, and \emph{Piecewise Convolution Neural Networks (PCNNs)}. The function of \emph{Vector Representation} is to transform words into low-dimensional vectors. The function of \emph{Linear Attenuation Simulation} is to assign weights to words. \emph{PCNNs} is used to extract feature vector of the sentence. The Attention Module is used to compute the weights of all sentences in a bag, and feed the bag features into a softmax classifier. And the Attention Module is comprised of \emph{Non-IID Relevance Embedding} and \emph{Classifying}. We elaborate on these parts in following paragraphs.

%Linear attenuation simulation is in the PCNNs Module and non-IID relevance embedding is in the Attention Module. This section contains five main parts: Vector Representation, Linear Attenuation Simulation, Piecewise convolution neural networks model (PCNNs Model), Non-IID Relevance Embedding, Classifying. We elaborate on these parts in following paragraphs.

\begin{figure*}[htb]
	\center
	\includegraphics[scale=0.25]{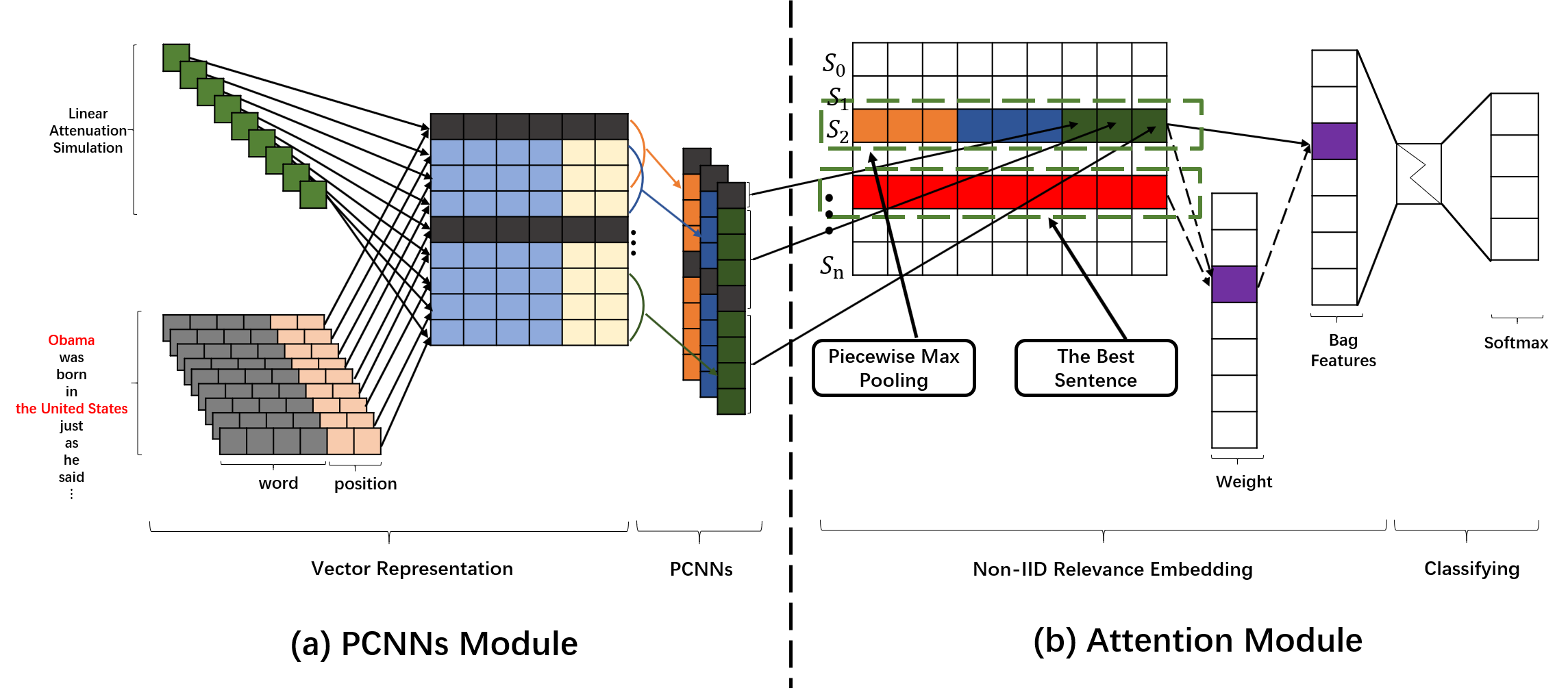}
	%\caption{The architecture of model used for distant supervision, showing how to extract the features of one sentence and predict the relation between entity pairs. The new matrix of input is coped with position-probability and vector representation.}
	\caption{The architecture of model. The red segment is the best sentence which can express the relation of r.}
\end{figure*}

\subsection{Vector Representation}
When using relation extraction, we require to translate each word to a low-dimensional vector. In this paper, we translate words into vectors by looking up the pre-trained word embeddings. In addition, position features (PFs) are used to specify entity pairs, which are also transformed into vectors by looking up the position embeddings.
\subsubsection{Word Embeddings:}Word embeddings are language modeling and feature learning techniques in NLP that map each word or phrase to a real-valued vector. They represent words between semantic and syntactic information. Given a sentence $\mathbf{X}= \{\mathbf{w_1},\mathbf{w_2},\dots,\mathbf{w_k}\}$, where each word $\mathbf{w_i}$ is represented by a real-valued vector. Word representations are encoded by vectors in an embedding matrix. In this paper, we use the Skip-gram model \cite{DBLP:journals/corr/abs-1301-3781} to train the word embeddings.

\subsubsection{Position Embeddings:}In distant supervised relation extraction, we focus on assigning labels to entity pairs. Similar to \cite{Zeng2014Relation}, we use position embeddings (PFs) to specify entity pairs. PFs are regarded as the combination of the relative distances from the current word to head entity and tail entity. For example, in the sentence ``\emph{Obama was born in the United States just as he has always said.}", the relative distances from ``\emph{he}" to head entity (\emph{Obama}) and tail entity (\emph{the United States}) are 7 and 3. Relative distances from ``\emph{in}" to head entity (\emph{Obama}) and tail entity (\emph{the United States}) are 4 and -1, respectively.

\begin{figure}[htb]
	\center
	\includegraphics[scale=0.4]{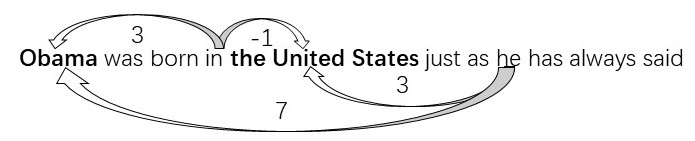}
    \caption{Position Embeddings}
\end{figure}

The position embedding matrices about entities are randomly initialized. Similar to the word embeddings, we transform the relative distances into real-valued vectors through looking up the position embedding matrices.

We assume that the size of word embedding is $d_w = 5$ and that the size of position embedding is $d_p = 1$. Finally, we combine the word embeddings and position embeddings of all words and transform it as a vector sequence $\mathbf{X}=\{\mathbf{w_1},\mathbf{w_2},\dots,\mathbf{w_k}\}$, where $k$ is the sentence length and $w_i \in\ R^d(d=d_w+d_p\ast2)$.

\subsection{Linear Attenuation Simulation}
In relation extraction, words which close to the target entities often contain more information about relations. On the contrary, when some words have long relative distances, these words are regarded as less or useless information about relations.

Suppose there is a sentence $\mathbf{X}$ consisting of $k$ words ($\mathbf{X}=\{\mathbf{w_1},\mathbf{w_2},\dots,\mathbf{w_k}\}$), containing a head entity and a tail entity. To exploit the information of all words, our model represents the sentence $\mathbf{X}$ with a real-valued matrix when predicting relation $r$. It is straightforward that the sentence is made up all words, $\{\mathbf{w_1},\mathbf{w_2},\dots,\mathbf{w_k}\}$. Each word contains different information which could decide relation of entity pairs. Then, the vector $\mathbf{X}$ is calculated as:
\begin{equation}
	\mathbf{X}=\{\gamma_1\mathbf{w_1},\gamma_2\mathbf{w_2},\dots,\gamma_i\mathbf{w_i},\dots,\gamma_k\mathbf{w_k}\}
\end{equation}
where $\gamma_i$ is the weights of each word. In general, we define $\gamma_i$ in two ways.

\subsubsection{Constant=1:}Normally, we think that each word in the sentence has the same weight to express the information of relation. We hence set $\gamma_i =\; 1$. Then, the sentence vector $\mathbf{X}$:
\begin{equation}
	\mathbf{X}=\{\mathbf{w_1},\mathbf{w_2},\dots,\mathbf{w_k}\}
\end{equation}

\subsubsection{Constant$\neq$1:}However, with the increase of sentence length, the weight continues to decrease about the relation. Therefore, if we regard each word as the same weight, the unimportant and the low-weight words will be equally computed with the high-weight words during the training and testing.

\begin{figure}[htb]
	\center
	\includegraphics[scale=0.4]{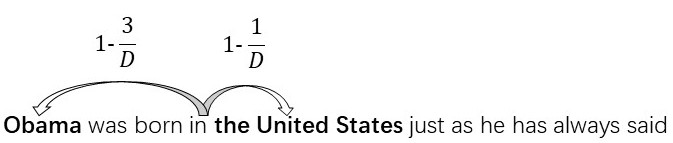}
    \caption{Linear Attenuation Simulation}
\end{figure}

So, we use linear attenuation simulation to reduce the impact of words with low weight. Hence, $\gamma_i$ is calculated as:
\begin{equation}
	\begin{array}{l}\gamma_i=\left\{\begin{array}{lc}(1-\frac{\vert d_{i1}\vert}D)+(1-\frac{\vert d_{i2}\vert}D)&if\;d_{ij}\leq D\\0\!&\!if\;d_{ij}>D\end{array}\right.\\\\\\\\\end{array}
\end{equation}
where $d_{i1}$ is referred as the relative distance about head entity. $d_{i2}$ is referred as the relative distance about tail entity. $j$ is the number which is 1 or 2. $D$ is referred as the threshold. If the distance of some words about entities is greater than $D$, their weights will be regarded as $0$.
Weights of ``\emph{in}" about ``\emph{Obama}" and ``\emph{the United States}" are $1-\frac3D$ and $1-\frac1D$. Thus, the weight of ``\emph{in}" is $2-\frac4D$.
Finally, we use the new $\mathbf{X}$ to accomplish the task of distant supervision.

\subsection{PCNNs}
In relation extraction, this model is employed to extract feature vectors of an instance.
\subsubsection{CNN:}Convolution neural networks is a typical neural networks. Convolution is an operation between the weight matrix $\mathbf{A}$, and the input matrix $\mathbf{B}$. $\mathbf{A}$ is regarded as the filter for the convolution. For example, we assume that $\mathbf{A}=(a_{ij})_{m\times n}$ and $\mathbf{B}= (b_{ij})_{m\times n}$, then $\mathbf{C}=\mathbf{A}\otimes\mathbf{B}={\textstyle\sum_{i=1}^m}{\textstyle\sum_{j=1}^n}a_{ij}b_{ij}$ is defined as convolution, where $\mathbf{C}$ is convolution, $m$ is the length of filter ($m=3$) and $n=d_w+d_p\ast2$. We consider $\mathbf{S}$ to be a sequence $\{\mathbf{q_1},\mathbf{q_2},\cdots,\mathbf{q_s}\}$. Normally, let $\mathbf{Q_{i:j}}$ refer to the concatenation of $\mathbf{q_i}$ to $\mathbf{q_j}$. Thus, the convolution operation between the matrix of sentence, $\mathbf{Q}$, and the matrix of weight, $\mathbf{W}$, results in another vector.
\begin{equation}
	\mathbf{c_j}=\mathbf{A}\otimes \mathbf{Q_{i:j}}
\end{equation}
where $j=i+m-1$.
\subsubsection{Piecewise Max-pooling:}PCNNs \cite{Zeng2015Distant}, a variation of CNN, adopts piecewise max-pooling in relation extraction to extract features. This method can obtain the structural information. Each convolution, $\mathbf{c_j}$, is divided into three parts $\mathbf{c_j}=\{\mathbf{c_{j1}},\mathbf{c_{j2}},\mathbf{c_{j3}}\}$ by head entity and tail entity. Then, the max-pooling procedure is performed in three parts separately. Next, we can concatenate all vectors $\mathbf{p_j}=\lbrack p_{j1},p_{j2},p_{j3}\rbrack$, which $p_{jh}=max\left(\mathbf{c_{jh}}\right)\left(h=1,2,3\right)$. Finally, we compute the feature vectors by a non-linear function at the output.

\subsection{Non-IID Relevance Embedding}
Given a bag $\mathbf{B}=\{\mathbf{s_0},\mathbf{s_1},\cdots,\mathbf{s_n}\}$, if we assume the predefined semantic relation is $r$, we can select the best sentence, $\mathbf{s_i}$, which can better perform the $r$ than the rest of sentences in the bag via multi-instance learning (MIL). And we consider that sentences in the bag can express $r$ and are non-IID. Traditionally, these sentences are often viewed as independent, which inevitably leads to loss of information for distant supervision.
To incorporate the non-IID, we compute similarity of remaining sentences with $\mathbf{s_i}$. There is a sentence, $\mathbf{s_j}$, in the bag. If $\mathbf{s_j}$ has a high similarity with $\mathbf{s_i}$, $\mathbf{s_j}$ could have a high weight in the bag. Higher similarity is, higher weight is in the bag. As shown in Figure 5, the bag has 4 sentences, and $\mathbf{s_1}$ is the best performance of $r$ by MIL. The weight of $\mathbf{s_2}$ about r can be computed by $\alpha_{1,2}$. In other words, $\alpha_{1,2}$ can select the weight of $\mathbf{s_2}$ about the relation of $r$. Hence, the weights of sentences about $r$ is calculated as:
\begin{equation}
\alpha_{i,j}=\frac{e_{i,j}}{{\displaystyle\sum_k}e_{i,k}}
\end{equation}
%where $\alpha_{i,j}$ is the weight of each sentence and $e_{i,j}$ is the weight of sentence about the $r$. $e_{i,j}$ is defined as:
where $\alpha_{i,j}$ is the weight of each sentence and $e_{i,j}$ is the similarity of sentence about the $r$. $e_{i,j}$ is calculated as:
\begin{equation}
	e_{i,j}=\frac{\mathbf{s_i}\cdot \mathbf{s_j}}{\vert\vert \mathbf{s_i}\vert\vert\times\vert\vert \mathbf{s_j}\vert\vert}
\end{equation}
where $\mathbf{s_i}$ is the best sentence of $r$, and $\mathbf{s_j}$ is sentence in the bag. The set vector $\mathbf{B}$ is calculated as a weighted sum of these sentence vectors:
\begin{equation}
	\mathbf{B}=\sum_j\alpha_{i,j}\mathbf{s_j}
\end{equation}

\begin{figure}[htb]
	\center
	\includegraphics[scale=0.25]{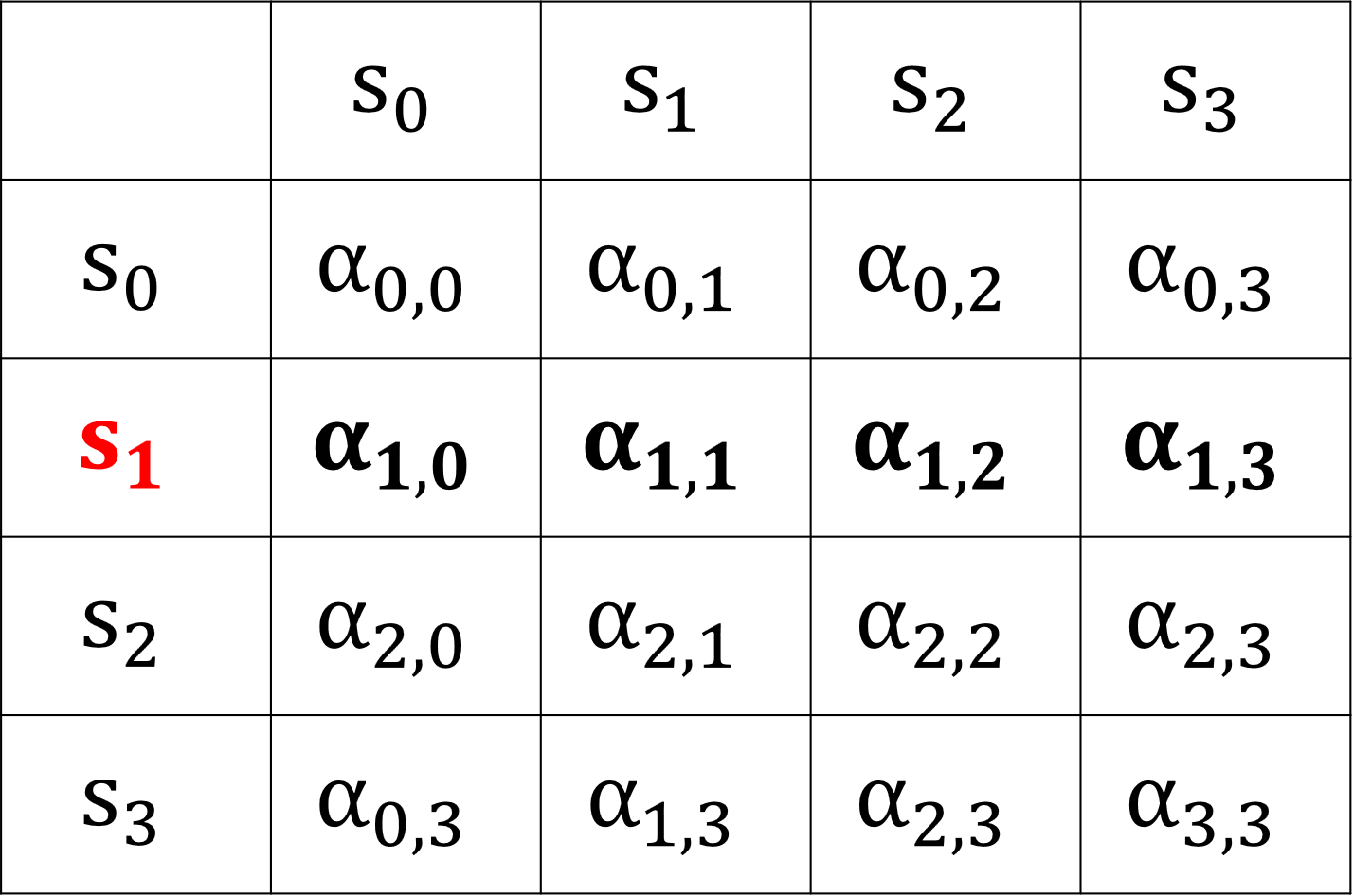}
	\caption{Non-IID Relevance Embedding. The part of bold font is the weights of sentences in a bag.}
\end{figure}

\subsection{Classifying}
In this section, we use softmax to get the conditional probability, as:
\begin{equation}
	p(r\vert B,\theta) =\frac{exp(o_r)}{{\displaystyle\sum_{k=1}}exp(o_k)}
\end{equation}
where $r$ is the representation of relation $r$, and $o$ is the final output, which is defined as:
\begin{equation}
	\mathbf{o}=\mathbf{M}\mathbf{B}+\mathbf{D}
\end{equation}
where M is the matrix of relations and D is a bias vector.
We define the objection function using cross-entropy\cite{Shore1980Axiomatic} as:
\begin{equation}
	J(\theta)=\sum_{i=1}^n\log\left(p\left(r_i\vert B_i,\theta\right)\right)
\end{equation}
where $n$ is the number of sentences and $\theta$ indicates all parameters of our model. In this paper, we combine dropout to prevent overfitting.

\section{Experiments}
Our experiments are intended to show that our model can capture high weight words and take full advantage of informative sentences for distant supervised relation extraction. In the experiments, we first introduce the dataset and evaluation metrics used. Next, we determine some parameters of our model by cross-validation. Finally, we evaluate the effects of linear attenuation simulation and non-IID relevance embedding, and we also compare our method to some classical methods.

\subsection{Dataset and Evaluation Metrics}
We evaluate our model on the New York Times (NYT)\footnote{http://iesl.cs.umass.edu/riedel/ecml.} corpus which is developed by \cite{Riedel2010Modeling} and has also been used by \cite{Hoffmann2011Knowledge,DBLP:conf/emnlp/SurdeanuTNM12,Lin2016Neural}. This dataset was generated by aligning Freebase relations. The sentences from 2005 to 2006 are used for training, and the sentences from 2007 are used for testing.

Following the previous work \cite{Lin2016Neural,Ji2017Distant}, we evaluate our method in the held-out evaluation. It evaluates our model by comparing the relation facts discovered from the test articles with those in Freebase. In the experiments, we assume that the NYT has the similar data structure every year. So, the held-out evaluation provides an approximate measure of precision without consuming human evaluation.  We report both the precision/recall curves and Precision@N (P@N) in our experiments.
\subsection{Experimental Settings}
\subsubsection{Word Embedding:}
In this paper, we employ the Skip-gram model\footnote{http://code.google.com/p/word2vec/} \cite{DBLP:journals/corr/abs-1301-3781} to train the word embeddings on the NYT corpus. The vector representations of words which learned by word2vec models have been shown to carry semantic meanings and are useful in NLP tasks.

\subsubsection{Parameters Setting:}
In this section, we study the influence of one parameter on our model: the threshold value $D$ is defined in Equation (3). We tune our models using three-fold validation on the training set. We use a grid search to determine the optional parameter: $D\in\{30,40,50,60,70,80\}$. For other parameters, we follow the settings used in \cite{Lin2016Neural}. For training, we set the iteration number over all the training data as 14. Table 1 shows all parameters used in the experiments.
\begin{table}[htb]
\centering
\label{my-label}
\scalebox{1}{
\begin{tabular}{c|c }
\hline \bf Setting & \bf Number \\ \hline
Window size         & 3      \\
Feature maps        & 230    \\
Word dimension      & 50     \\
Position dimension  & 5      \\
Batch size          & 160    \\
Learning rate       & 0.01   \\
Dropout probability & 0.5    \\
Threshold           & 60      \\  \hline
\end{tabular}}
\centering
\caption{\label{font-table} Parameters Setting}
\end{table}

\begin{table*}[ht]
 \scalebox{1.1}{
\begin{tabular}{c|c|c|c|c|c}
\hline
P@N(\%) & PCNNs+MIL & PCNNs+ATT & PCNNs+W & PCNNs+N & PCNNs+WN \\ \hline
P@100   & 72.3     & 76.2     &{\bf 83.0}    & 81.0    &{\bf 83.0}     \\ \hline
P@200   & 69.7     & 73.1     & 77.0    & 79.5    &{\bf 82.0}     \\ \hline
P@300   & 64.1     & 67.4     & 72.0    & 76.7    &{\bf 80.3}     \\ \hline
Average & 68.7     & 72.2     & 77.0    & 79.1    &{\bf 81.8}     \\ \hline
\end{tabular}}
\centering
\caption{P@N for relation extraction}
\end{table*}

\subsection{Effect of Linear Attenuation Simulation and Non-IID Relevance Embedding}
To prove the influence about linear attenuation simulation and non-IID relevance embedding, we compared with different methods by held-out evaluation. We select \emph{PCNNs+ATT} as our baseline. \emph{PCNNs} represents \emph{CNN} with piecewise max-pooling, and \emph{ATT} represents sentence-level attention. \emph{PCNNs+ATT} has better performance than other methods in distant supervision. In order to demonstrate the validity of our method, we carried out some experiments.
\emph{PCNNs+W} represents linear attenuation simulation with \emph{PCNNs}. \emph{PCNNs+N} represents non-IID relevance embedding with \emph{PCNNs}.
\emph{PCNNs+WN} represents linear attenuation simulation and non-IID relevance embedding with \emph{PCNNs}. To determine the threshold value, $D$, we select the different values in the experiments, $D\in\{30,40,50,60,70,80\}$. Experimental results are in Figure 6(b).

\begin{figure}[htbp]
   \centering
   \subfigure[Comparison of baseline and our approach.]{\includegraphics[width=0.45\textwidth]{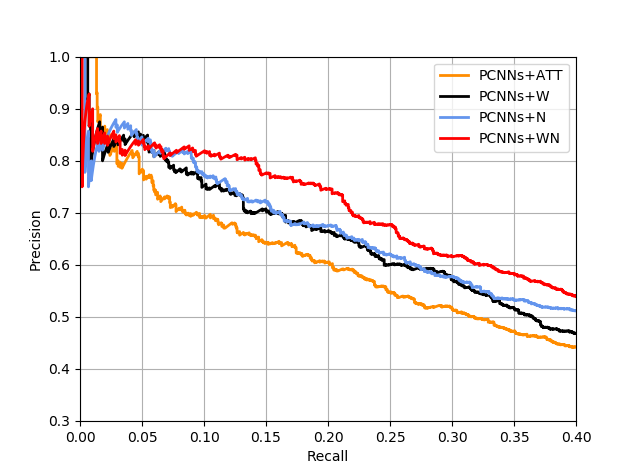}}
   \hspace{-0.0cm}
   \subfigure[Comparison of different values (D).]{\includegraphics[width=0.45\textwidth]{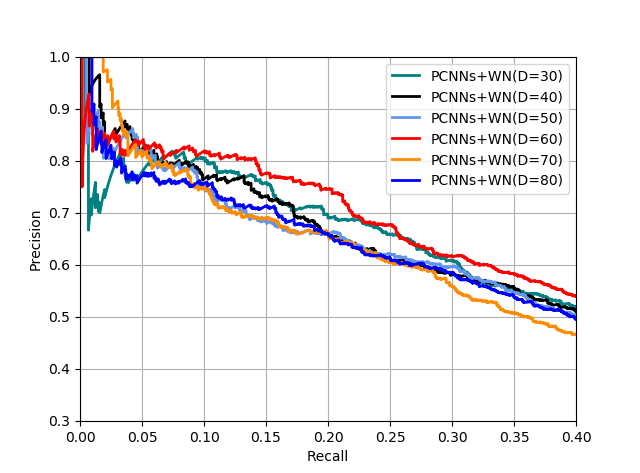}}
   \vspace{-0.3cm}
   \caption{Effect of Linear Attenuation Simulation and Non-IID Relevance Embedding.}
   \vspace{0.0cm}
   \label{fig:PPX_KL}
\end{figure}

Figure 6 shows that when linear attenuation simulation and non-IID relevance embedding is used in \emph{PCNNs}, our method has achieved good results in relation extraction. Figure 6(a) shows that when linear attenuation simulation or non-IID relevance embedding is used alone in \emph{PCNNs}, they all perform better than \emph{PCNNs+ATT}. And \emph{PCNNs+WN} achieves the highest precision compared to other methods. These results indicate that linear attenuation simulation can selectively assign different weights to words, and alleviate wrong labels for relation extraction. Moreover, we also notice that non-IID relevance embedding can capture the relevance of sentences, and enhance the correct labels.
Figure 6(b) shows that when $D$ is 60, our method can get the best performance. Hence, linear attenuation simulation and non-IID relevance embedding are important factors in distant supervision.

\subsection{Comparison with Traditional Approaches}
\subsubsection{Held-out Evaluation:}
To evaluate the proposed method, we select the following seven traditional methods for comparison.
\begin{itemize}
\item \textbf{Mintz} \cite{Mintz2009Distant} proposed a traditional distant supervision model.

\item \textbf{MultiR} \cite{Hoffmann2011Knowledge} proposed a probabilistic graphical model with multi-instance learning.

\item \textbf{MIML} \cite{DBLP:conf/emnlp/SurdeanuTNM12} proposed a multi-instance and multi-label model.

\item \textbf{PCNNs+MIL} \cite{Zeng2015Distant} proposed piecewise convolutional neural networks (PCNNs) with multi-instance learning.

\item \textbf{PCNNs+ATT} \cite{Lin2016Neural} proposed a selective attention over instances with PCNNs and CNNs.

\item \textbf{APCNNs+D} \cite{Ji2017Distant} proposed background information of entities by an attention layer to help relation classification.

\item \textbf{SEE-TRANS} \cite{DBLP:HeCLZZZ18} proposed syntax-aware entity embedding with PCNNs+ATT.

\item \textbf{PCNNs+WN} is our method with PCNNs.
\end{itemize}

\begin{figure}[htb]
	\center
	\includegraphics[scale=0.45]{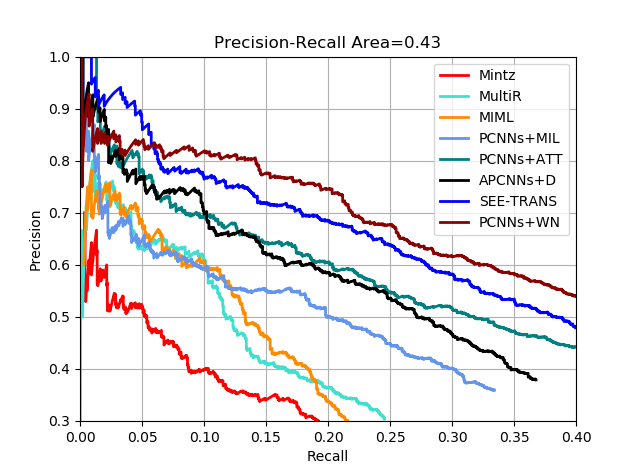}
	\caption{Performance comparison among different methods}
\end{figure}

Figure 7 shows that the precision-recall curves for each method. We can observe that:  (1) \emph{PCNNs+WN} achieves higher precision. \emph{PCNNs+WN} enhance the mean average precision to approximately $43\%$. When the recall is greater than 0.07, performance of our method drops out quickly. The results demonstrate that our method is an effective way to distant supervised relation extraction and \emph{PCNNs+WN} can alleviate the error propagation. (2) The precision of our method has declined when recall is less than 0.07. Because linear attenuation simulation reduces some words in long sentences. Maybe these words have effects on certain performance of relations.
But in the experiments, our method has better performance than other methods and improves the overall effect of relation extraction. These results demonstrate that our method possesses important effects for distant supervision.

\begin{figure*}[htb]
	\center
	\includegraphics[scale=0.35]{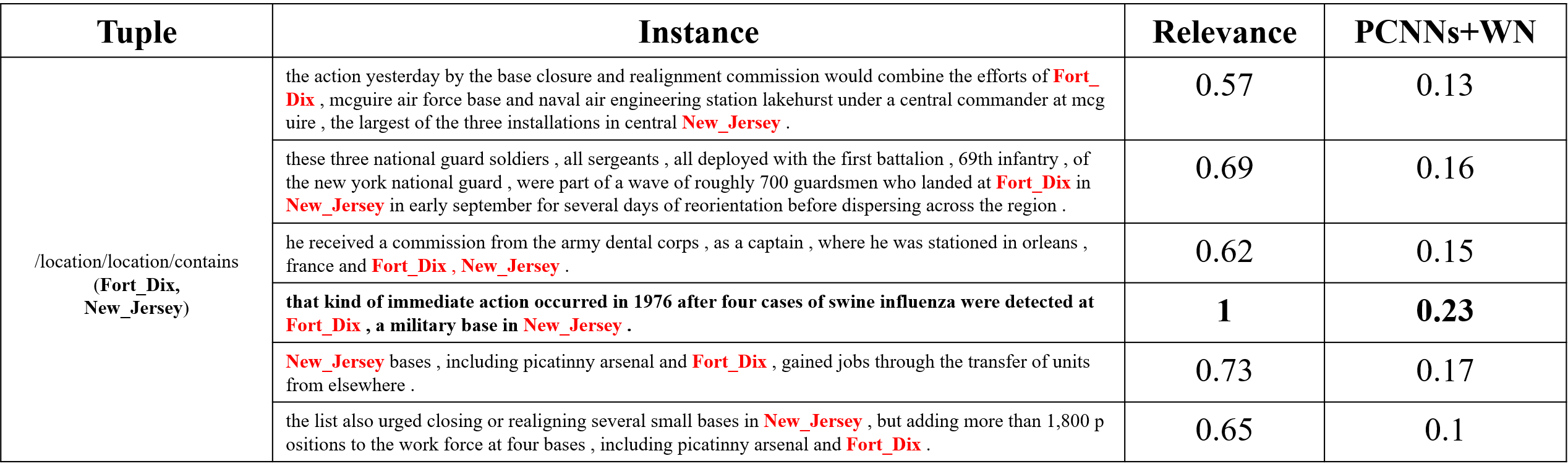}
	\caption{Some examples of PCNNs+WN in the NYT}
\end{figure*}

\subsubsection{P@N Evaluation:}In this section, we report the P@100, P@200, P@300 and the average of them for \emph{PCNNs+MIL}, \emph{PCNNs+ATT}, \emph{PCNNs+W}, \emph{PCNNs+W}, and \emph{PCNNs+WN}.
%\emph{PCNNs+AVE} \citep{Lin2016Neural} represents each sentence set as the average vector of sentences inside the set.

Table 2 shows that:  (1) \emph{PCNNs+WN} achieves the best performance in all test settings. \emph{PCNNs+WN} outperforms \emph{PCNNs+ATT} over $9.6\%$ in the average. It demonstrates the validity of linear attenuation simulation and non-IID relevance embedding for distant supervision. (2) For both \emph{PCNNs+W} and \emph{PCNNs+N}, the results of these methods are better than \emph{PCNNs+ATT}. Because linear attenuation simulation can alleviate words of low weight and non-IID relevance embedding can capture valid information of each sentence about relation in a bag.

\subsection{Case Study}
Figure 8 shows an example of \emph{PCNNs+WN} from the testing data. The entity-relation tuple is (\emph{Fort-Dix}, \emph{New-Jersey}, \emph{contains}). There are 6 sentences containing the entity pair. The 4-th sentence, being the part of bold font, is the best sentence to express ``\emph{contains}". Our model not only can capture relation of sentences, but also can analyze correlations between 4-th sentence and each sentence in this bag. Relevance represents the correlation between 4-th sentence and each sentence in a bag.
Hence, our model assigns high weights to valid sentences for our task. We argue that linear attenuation simulation and non-IID relevance embedding can enhance the performance in distant supervision. We can clearly distinguish valid sentences and invalid sentences. Therefore, linear attenuation simulation and non-IID relevance embedding can provide more information of sentences and alleviate wrong labels.

\section{Related Work}

Relation extraction is one of the most important tasks in NLP. Many methods have been proposed in relation extraction, such as bootstrapping, unsupervised relation discovery and supervised classification. Supervised methods are the classical approaches to deal with the relation extraction and perform good expression \cite{Bunescu2005Subsequence,Zhang2006Adapting,Zelenko2003Kernel}. However, these approaches heavily depend on high quality training data.

Recently, deep learning has been widely used to automatically extract relation. It is the most representative progress in deep neural networks to cope with relation extraction, such as convolutional neural network (CNN) \cite{Zeng2014Relation,Santos2015Classifying}, recurrent neural networks (RNN) \cite{Cho2014Learning,DBLP:conf/acl/LiuWLJZW15}, long short-term memory network (LSTM) \cite{Miwa2016End,Yan2015Classifying,Sundermeyer2012LSTM} and attention-based bidirectional LSTM \cite{Zhou2016Attention}.
In general, relation extraction need mass high quality training data, which could spend much time and energy. To figure out this issue, \cite{Mintz2009Distant} used distant supervision to automatically produce training data via aligning KBs and texts. They assume that if two entities have a relation in KBs, all sentences which contain these two entities will express the same relation. Distant supervision is an effective method to automatically label datasets, but it often suffers from incorrect information. To alleviate this issue, some researchers regarded relation classification as a multi-instance multi-label learning problem \cite{Riedel2010Modeling,Hoffmann2011Knowledge,Sundermeyer2012LSTM}. The term ‘multi-instance learning’ was proposed to predict the drug activity \cite{DBLP:journals/ai/DietterichLL97}. In multi-instance learning, the uncertainty sentences can be regarded as the label of bag. Thus, the focus of multi-instance learning is to discriminate the label of bag. However, multi-instance learning is difficult to apply in neural network models. \cite{Zeng2015Distant} proposed at-least-one multi-instance learning and piecewise convolutional neural networks(\emph{PCNNs+MIL}) to extract the relations in distant supervision.
But \emph{PCNNs+MIL} ignores a lot of useful information. To capture the informative sentences and reduce the influence of wrong labelled sentences, a sentence-level attention mechanism over multiple instances was proposed \cite{Lin2016Neural,Ji2017Distant,Liu2017A}. To exploit impact between syntax information and relation extraction, \cite{DBLP:HeCLZZZ18} proposed to learn syntax-aware entity embedding for relation extraction. Learning from non-IID data is a recent topic \cite{Cao2014Non,DBLP:conf/aaai/ShiLGCS17,DBLP:conf/ijcai/PangCCL17} to address the intrinsic data complexities, with preliminary work reported such as for clustering \cite{Wang2011Coupled}. However, the non-IID in distant supervision is seldom exploited.

Traditional methods assume that each word of the sentence is regarded as the same weight and each sentence are independent in a bag. Actually, each word could not have the same weight in the sentence and each sentence are not independent in a bag. To address these issues, we propose a novel model which can capture informative words and sentences.

\section{Conclusion}

In this paper, we exploit linear attenuation simulation and non-IID relevance embedding with piecewise convolutional neural networks (PCNNs) for distant supervised relation extraction. We apply the linear attenuation simulation to capture the words of high weights in the sentence, and then we use the non-IID relevance embedding to extract connections about surrounding sentences in the bag. We conduct experiments on a widely used benchmark dataset. The experiments show that proposed method has better performance than comparable methods. These results demonstrate that our approach can effectively deal with the task of relation extraction.

In the future, we will explore the following directions:

\begin{itemize}
\item Our method not only can be used in distant supervised relation extraction, but also can be used in other fields, such as event detection and question answering.
\item Reinforcement learning (RL) is one of the effective methods for NLP task. In the future, we can combine our method with reinforcement learning for distant supervision.
\end{itemize}

\section{Acknowledgements}
We would like to thank Yuxiang Zhou, Rihai Su, Qian Liu and Luyang Liu for their insightful comments and suggestions. We also very appreciate the comments from anonymous reviewers which will help further improve our work. This work is supported by National Key R\&D Plan(No.2017YFB0803302), National Natural Science Foundation of China (No.61751201) and Research Foundation of Beijing Municipal Science \& Technology  Commission (Grant No. Z181100008918002).

\bibliographystyle{aaai}
\bibliography{refer}

\begin{thebibliography}{}

\bibitem[\protect\citeauthoryear{Bunescu and
  Mooney}{2005}]{Bunescu2005Subsequence}
Bunescu, R.~C., and Mooney, R.~J.
\newblock 2005.
\newblock Subsequence kernels for relation extraction.
\newblock In {\em Proceedings of NIPS},  171--178.

\bibitem[\protect\citeauthoryear{Cao}{2014}]{Cao2014Non}
Cao, L.
\newblock 2014.
\newblock Non-iidness learning in behavioral and social data.
\newblock {\em The Computer Journal} 57(9):1358--1370.

\bibitem[\protect\citeauthoryear{Cho \bgroup et al\mbox.\egroup
  }{2014}]{Cho2014Learning}
Cho, K.; Van~Merrienboer, B.; Gulcehre, C.; Bahdanau, D.; Bougares, F.;
  Schwenk, H.; and Bengio, Y.
\newblock 2014.
\newblock Learning phrase representations using rnn encoder-decoder for
  statistical machine translation.
\newblock {\em Computer Science}.

\bibitem[\protect\citeauthoryear{Dietterich, Lathrop, and
  Lozano{-}P{\'{e}}rez}{1997}]{DBLP:journals/ai/DietterichLL97}
Dietterich, T.~G.; Lathrop, R.~H.; and Lozano{-}P{\'{e}}rez, T.
\newblock 1997.
\newblock Solving the multiple instance problem with axis-parallel rectangles.
\newblock {\em Artif. Intell.} 89(1-2):31--71.

\bibitem[\protect\citeauthoryear{He \bgroup et al\mbox.\egroup
  }{2018}]{DBLP:HeCLZZZ18}
He, Z.; Chen, W.; Li, Z.; Zhang, M.; Zhang, W.; and Zhang, M.
\newblock 2018.
\newblock {SEE:} syntax-aware entity embedding for neural relation extraction.
\newblock In {\em Proceedings of AAAI}.

\bibitem[\protect\citeauthoryear{Hoffmann \bgroup et al\mbox.\egroup
  }{2011}]{Hoffmann2011Knowledge}
Hoffmann, R.; Zhang, C.; Ling, X.; Zettlemoyer, L.; and Weld, D.~S.
\newblock 2011.
\newblock Knowledge-based weak supervision for information extraction of
  overlapping relations.
\newblock In {\em Proceedings of ACL},  541--550.

\bibitem[\protect\citeauthoryear{Huynh \bgroup et al\mbox.\egroup
  }{2011}]{DBLP:conf/acsc/HuynhTMS11}
Huynh, D.; Tran, D.; Ma, W.; and Sharma, D.
\newblock 2011.
\newblock A new term ranking method based on relation extraction and graph
  model for text classification.
\newblock In {\em Proceedings of ACSC},  145--152.

\bibitem[\protect\citeauthoryear{Ji \bgroup et al\mbox.\egroup
  }{2017}]{Ji2017Distant}
Ji, G.; Liu, K.; He, S.; and Zhao, J.
\newblock 2017.
\newblock Distant supervision for relation extraction with sentence-level
  attention and entity descriptions.
\newblock In {\em Proceedings of AAAI},  3060--3066.

\bibitem[\protect\citeauthoryear{Lin \bgroup et al\mbox.\egroup
  }{2016}]{Lin2016Neural}
Lin, Y.; Shen, S.; Liu, Z.; Luan, H.; and Sun, M.
\newblock 2016.
\newblock Neural relation extraction with selective attention over instances.
\newblock In {\em Proceedings of ACL},  2124--2133.

\bibitem[\protect\citeauthoryear{Liu \bgroup et al\mbox.\egroup
  }{2015}]{DBLP:conf/acl/LiuWLJZW15}
Liu, Y.; Wei, F.; Li, S.; Ji, H.; Zhou, M.; and Wang, H.
\newblock 2015.
\newblock A dependency-based neural network for relation classification.
\newblock In {\em Proceedings of ACL},  285--290.

\bibitem[\protect\citeauthoryear{Liu \bgroup et al\mbox.\egroup
  }{2017}]{Liu2017A}
Liu, T.; Wang, K.; Chang, B.; and Sui, Z.
\newblock 2017.
\newblock A soft-label method for noise-tolerant distantly supervised relation
  extraction.
\newblock In {\em Proceedings of EMNLP},  1790--1795.

\bibitem[\protect\citeauthoryear{Mcdonald and
  Nivre}{2007}]{Mcdonald2007Characterizing}
Mcdonald, R.~T., and Nivre, J.
\newblock 2007.
\newblock Characterizing the errors of data-driven dependency parsing models.
\newblock In {\em Proceedings of EMNLP-CoNLL},  122--131.

\bibitem[\protect\citeauthoryear{Mikolov \bgroup et al\mbox.\egroup
  }{2013}]{DBLP:journals/corr/abs-1301-3781}
Mikolov, T.; Chen, K.; Corrado, G.; and Dean, J.
\newblock 2013.
\newblock Efficient estimation of word representations in vector space.
\newblock {\em CoRR} abs/1301.3781.

\bibitem[\protect\citeauthoryear{Mintz \bgroup et al\mbox.\egroup
  }{2009}]{Mintz2009Distant}
Mintz; Mike; Steven; Jurafsky; and Dan.
\newblock 2009.
\newblock Distant supervision for relation extraction without labeled data.
\newblock In {\em Proceedings of ACL-IJCNLP},  1003--1011.

\bibitem[\protect\citeauthoryear{Miwa and Bansal}{2016}]{Miwa2016End}
Miwa, M., and Bansal, M.
\newblock 2016.
\newblock End-to-end relation extraction using lstms on sequences and tree
  structures.
\newblock In {\em Proceedings of ACL},  1105--1116.

\bibitem[\protect\citeauthoryear{Pang \bgroup et al\mbox.\egroup
  }{2017}]{DBLP:conf/ijcai/PangCCL17}
Pang, G.; Cao, L.; Chen, L.; and Liu, H.
\newblock 2017.
\newblock Learning homophily couplings from non-iid data for joint feature
  selection and noise-resilient outlier detection.
\newblock In {\em Proceedings of IJCAI},  2585--2591.

\bibitem[\protect\citeauthoryear{Riedel, Yao, and
  McCallum}{2010}]{Riedel2010Modeling}
Riedel, S.; Yao, L.; and McCallum, A.
\newblock 2010.
\newblock Modeling relations and their mentions without labeled text.
\newblock In {\em Proceedings of ECML/PKDD},  148--163.

\bibitem[\protect\citeauthoryear{Sadeghi, Divvala, and
  Farhadi}{2015}]{DBLP:conf/cvpr/SadeghiDF15}
Sadeghi, F.; Divvala, S.~K.; and Farhadi, A.
\newblock 2015.
\newblock Viske: Visual knowledge extraction and question answering by visual
  verification of relation phrases.
\newblock In {\em Proceedings of CVPR},  1456--1464.

\bibitem[\protect\citeauthoryear{Santos, Xiang, and
  Zhou}{2015}]{Santos2015Classifying}
Santos, C. N.~D.; Xiang, B.; and Zhou, B.
\newblock 2015.
\newblock Classifying relations by ranking with convolutional neural networks.
\newblock {\em Computer Science} 86(86):132--137.

\bibitem[\protect\citeauthoryear{Shi \bgroup et al\mbox.\egroup
  }{2017}]{DBLP:conf/aaai/ShiLGCS17}
Shi, Y.; Li, W.; Gao, Y.; Cao, L.; and Shen, D.
\newblock 2017.
\newblock Beyond {IID:} learning to combine non-iid metrics for vision tasks.
\newblock In {\em Proceedings of AAAI},  1524--1531.

\bibitem[\protect\citeauthoryear{Shore and Johnson}{1980}]{Shore1980Axiomatic}
Shore, J.~E., and Johnson, R.~W.
\newblock 1980.
\newblock Axiomatic derivation of the principle of maximum entropy and the
  principle of minimum cross-entropy.
\newblock {\em Information Theory IEEE Transactions on} 26(1):26--37.

\bibitem[\protect\citeauthoryear{Sundermeyer, Schluter, and
  Ney}{2012}]{Sundermeyer2012LSTM}
Sundermeyer, M.; Schluter, R.; and Ney, H.
\newblock 2012.
\newblock Lstm neural networks for language modeling.
\newblock In {\em Proceedings of INTERSPEECH},  601--608.

\bibitem[\protect\citeauthoryear{Surdeanu \bgroup et al\mbox.\egroup
  }{2012}]{DBLP:conf/emnlp/SurdeanuTNM12}
Surdeanu, M.; Tibshirani, J.; Nallapati, R.; and Manning, C.~D.
\newblock 2012.
\newblock Multi-instance multi-label learning for relation extraction.
\newblock In {\em Proceedings of EMNLP-CoNLL},  455--465.

\bibitem[\protect\citeauthoryear{Wang \bgroup et al\mbox.\egroup
  }{2011}]{Wang2011Coupled}
Wang, C.; Cao, L.; Wang, M.; Li, J.; Wei, W.; and Ou, Y.
\newblock 2011.
\newblock Coupled nominal similarity in unsupervised learning.
\newblock In {\em Proceedings of CIKM},  973--978.

\bibitem[\protect\citeauthoryear{Yan \bgroup et al\mbox.\egroup
  }{2009}]{DBLP:conf/acl/YanOMYI09}
Yan, Y.; Okazaki, N.; Matsuo, Y.; Yang, Z.; and Ishizuka, M.
\newblock 2009.
\newblock Unsupervised relation extraction by mining wikipedia texts using
  information from the web.
\newblock In {\em Proceedings of ACL/IJCNLP},  1021--1029.

\bibitem[\protect\citeauthoryear{Yan \bgroup et al\mbox.\egroup
  }{2015}]{Yan2015Classifying}
Yan, X.; Mou, L.; Li, G.; Chen, Y.; Peng, H.; and Jin, Z.
\newblock 2015.
\newblock Classifying relations via long short term memory networks along
  shortest dependency path.
\newblock {\em Computer Science} 42(1):56--61.

\bibitem[\protect\citeauthoryear{Zelenko, Aone, and
  Richardella}{2003}]{Zelenko2003Kernel}
Zelenko, D.; Aone, C.; and Richardella, A.
\newblock 2003.
\newblock Kernel methods for relation extraction.
\newblock {\em Journal of Machine Learning Research} 3(3):1083--1106.

\bibitem[\protect\citeauthoryear{Zeng \bgroup et al\mbox.\egroup
  }{2014}]{Zeng2014Relation}
Zeng, D.; Liu, K.; Lai, S.; Zhou, G.; and Zhao, J.
\newblock 2014.
\newblock Relation classification via convolutional deep neural network.
\newblock In {\em Proceedings of COLING},  2335--2344.

\bibitem[\protect\citeauthoryear{Zeng \bgroup et al\mbox.\egroup
  }{2015}]{Zeng2015Distant}
Zeng, D.; Liu, K.; Chen, Y.; and Zhao, J.
\newblock 2015.
\newblock Distant supervision for relation extraction via piecewise
  convolutional neural networks.
\newblock In {\em Proceedings of EMNLP},  1753--1762.

\bibitem[\protect\citeauthoryear{Zhang and Zhou}{2006}]{Zhang2006Adapting}
Zhang, M.~L., and Zhou, Z.~H.
\newblock 2006.
\newblock Adapting rbf neural networks to multi-instance learning.
\newblock {\em Neural Proceedings Letters} 23(1):1--26.

\bibitem[\protect\citeauthoryear{Zhou \bgroup et al\mbox.\egroup
  }{2016}]{Zhou2016Attention}
Zhou, P.; Shi, W.; Tian, J.; Qi, Z.; Li, B.; Hao, H.; and Xu, B.
\newblock 2016.
\newblock Attention-based bidirectional long short-term memory networks for
  relation classification.
\newblock In {\em Proceedings of ACL},  207--212.

\end{thebibliography}
\end{document}